\documentclass[10pt,twocolumn,letterpaper]{article}

\usepackage{cvpr}              % To produce the CAMERA-READY version
\usepackage{float}

\usepackage[dvipsnames]{xcolor}

\definecolor{cvprblue}{rgb}{0.21,0.49,0.74}
\usepackage[pagebackref,breaklinks,colorlinks,citecolor=cvprblue]{hyperref}
\usepackage{multirow}
\usepackage{xcolor}

\newcommand{\myparagraph}[1]{\vspace{2pt}\noindent{\bf #1}}

\def\paperID{22} % *** Enter the Paper ID here
\def\confName{CVPR}
\def\confYear{2024}

\begin{document}

\title{\vspace{-6pt}Toward a Diffusion-Based Generalist for Dense Vision Tasks\vspace{-6pt}}

%%%%%%%%% AUTHORS - PLEASE UPDATE
\author{
Yue Fan$^{1,2}$\thanks{\scriptsize{Intern at
Google during the project.}}, Yongqin Xian$^2$, Xiaohua Zhai$^3$, Alexander Kolesnikov$^3$, \\Muhammad Ferjad Naeem$^4$, Bernt Schiele$^1$, Federico Tombari$^{2,5}$ \\
\small{$^{1}$Max Planck Institute for Informatics, Saarland Informatics,}\\
\small{$^{2}$Google,} 
\small{$^{3}$Google DeepMind, $^{4}$ETH Zurich, $^{5}$TU Munich}
}

\makeatletter
\let\@oldmaketitle\@maketitle%
\renewcommand{\@maketitle}{\@oldmaketitle%
    % \vspace{-2em}
    \centering
    \includegraphics[width=.85\linewidth]{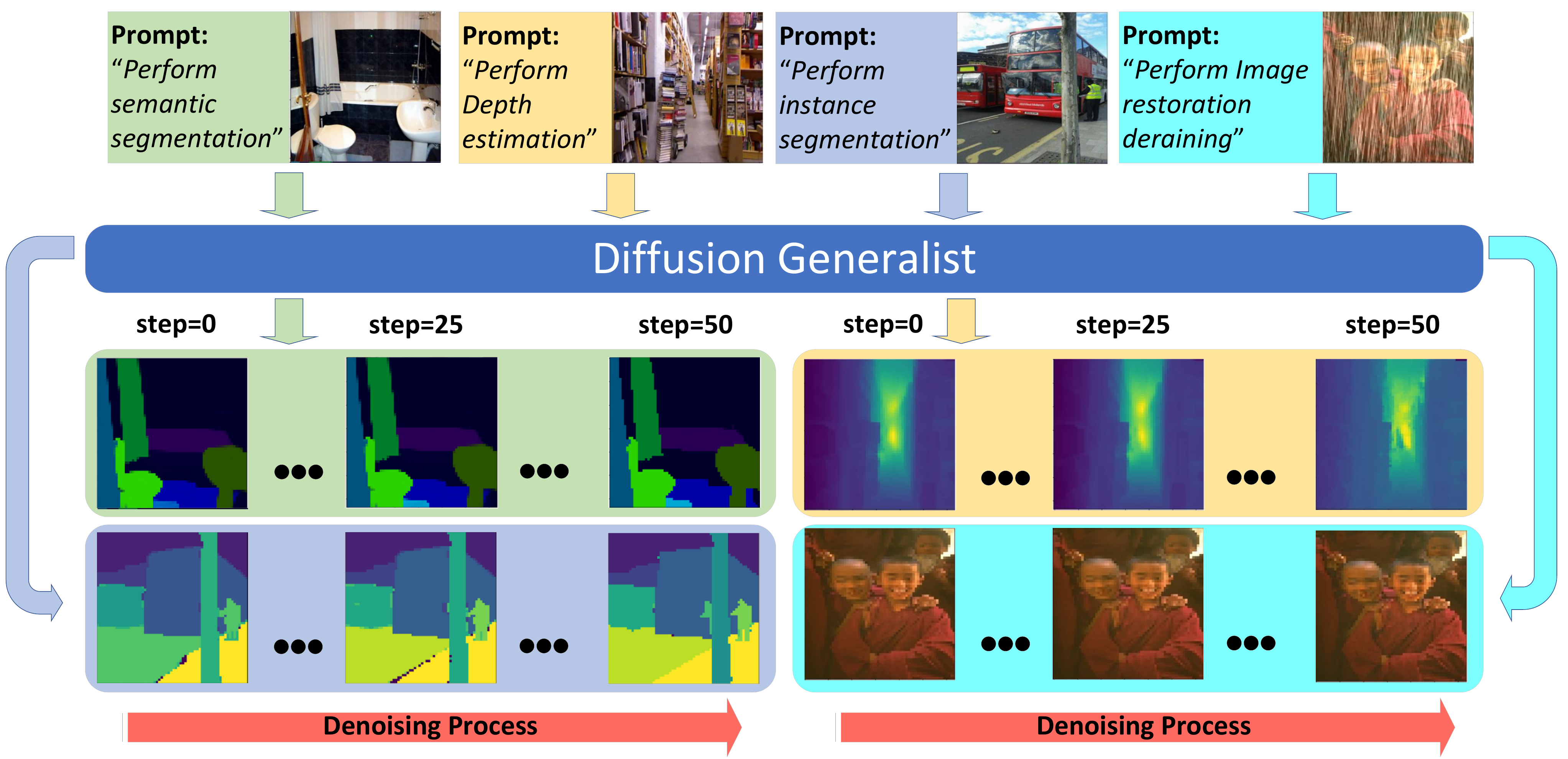}
    \captionof{figure}{We present a diffusion-based vision generalist for dense vision tasks. Given an input image, the model performs the corresponding task following the text instruction. We showcase the effectiveness of our model on depth estimation, semantic segmentation, panoptic segmentation, and three types of image restoration tasks. The images are the actual output of our model.
    }
    % \vspace{-5pt}
    \label{fig:teaser}
   \bigskip}                   %
\makeatother

\maketitle

\begin{abstract}
Building generalized models that can solve many computer vision tasks simultaneously is an intriguing direction.
% The current de facto approach for these tasks is to design model architectures and loss functions that are tailored to the task at hand. 
Recent works have shown image itself can be used as a natural interface for general-purpose visual perception and demonstrated inspiring results.
In this paper, we explore diffusion-based vision generalists, where we unify different types of dense prediction tasks as conditional image generation and re-purpose pre-trained diffusion models for it.
However, directly applying off-the-shelf latent diffusion models leads to a quantization issue.
Thus, we propose to perform diffusion in pixel space and provide a recipe for finetuning pre-trained text-to-image diffusion models for dense vision tasks.  
% For example, we found that extracting features from the original image is better than just feeding the whole image into the diffusion models.
In experiments, we evaluate our method on four different types of tasks and show competitive performance to the other vision generalists. 
\end{abstract}

\section{Introduction} \label{sec:intro}
% \begin{figure}[H]
% \centering
% \includegraphics[width=0.9\textwidth]{teaser.pdf}
% \caption{
% teaser.
% }
% \label{fig:teaser}
% \end{figure}

% Inspired by the prominent unified approach in natural language processing (NLP) tasks, there has been a growing interest in developing unified approaches for vision tasks.
% However, due to the intrinsic difference between the output format of different vision tasks, unifying different vision tasks is very difficult.

The field of artificial intelligence has made significant progress in building generalized model frameworks.
In particular, autoregressive transformers \cite{vaswani2017attention} have become a prominent unified approach in Natural Language Processing (NLP), effectively addressing a wide range of tasks with a singular model architecture \cite{touvron2023llama,devlin2018bert,radford2019gpt,raffel2020t5}.
However, in computer vision (CV), building a unified framework remains challenging due to the inherent diversity of the tasks and output formats.
% For example, segmentation and object detection.
Consequently, state-of-the-art computer vision models still have many complex task-specific designs \cite{carion2020detr,cheng2021maskformer,cheng2022mask2former,li2022binsformer,wang2022uformer}, making it difficult for feature sharing across tasks and, thus, limiting knowledge transfer. % across different vision tasks. % hamper positive knowledge transfer across different tasks.

% There has been an increasing interest in unified approaches for vision tasks.
The stark contrast between NLP and CV has given rise to a growing interest in developing unified approaches for vision tasks \cite{lu2022unified,chen2021pix2seq,chen2022pix2,wang2023painter,wang2023seggpt,zhu2022uni}. 
Recently, \cite{wang2023painter,wang2023seggpt} have shown image itself can be used as a robust interface for unifying different vision tasks and demonstrated good performance.
% Diffusion models \cite{???} have demonstrated remarkable capabilities of transforming diverse text prompts into realistic images, even for novel concepts.
In this paper, we propose a multi-task diffusion generalist for dense vision tasks by reformulating the dense prediction tasks as conditional image generation, and re-purpose pre-trained latent diffusion models for it. 
Fig. \ref{fig:teaser} visualizes the output of our model on semantic segmentation, panoptic segmentation, depth estimation, and image restoration.
Based on text prompts, our model can perform different tasks with one set of parameters.
% The model is trained jointly on a diverse set of tasks and the text prompts are used as explicit task indicators, guiding the generation process to produce the visual task output corresponding to the input image.
However, directly finetuning the pre-trained latent diffusion models (e.g. Stable Diffusion \cite{rombach2022sd}) leads to quantization errors for segmentation tasks (see Table \ref{tab:latent_vs_pixel}). 
% As shown in Table \ref{tab:latent_vs_pixel}, when mapping from the latent space to the pixel space, visually uniform regions actually have pixels of many different RGB values.
% This variance can lead to inaccurate class mappings, and consequently, suboptimal performance.
To this end, we propose to do pixel-space diffusion which effectively improves the generation quality and does not suffer from quantization errors. %when producing discrete colors.
Moreover, our exploration into training diffusion models as vision generalists reveals a list of interesting findings as follows:
\begin{itemize}
    \item Diffusion-based generalists show superior performance over the non-diffusion-based generalists on tasks involving semantics or global understanding of the scene.
    \item We find conditioning on the image feature extracted from powerful pre-trained image encoders results in better performance than directly conditioning on the raw image.
    \item Pixel diffusion is better than latent diffusion as it does not have the quantization issue while upsampling.
    \item We observe that text-to-image generation pre-training stabilizes the training and leads to better performance.
    % \item We find that diffusion-based multi-task generalists tend to suffer from underfitting and increasing the batch size can largely mitigate the issue.
\end{itemize}

% To summarize, we investigate a diffusion-based multi-task generalist for dense vision tasks by extending denoising diffusion probabilistic models.
In experiments, we demonstrate the model's versatility across six different dense prediction tasks on depth estimation, semantic segmentation, panoptic segmentation, image denoising, image draining, and light enhancement. Our method achieves competitive performance to the current state-of-the-art in many settings.

\section{Related Work}
\myparagraph{Unified framework \& Unified model:}
Efforts have been made to unify various vision tasks with a single model, resulting in several vision generalists \cite{lu2022unified,chen2021pix2seq,chen2022pix2,wang2023painter,wang2023seggpt,kolesnikov2022uvim}.
Inspired by the success of sequence-to-sequence modeling in Natural Language Processing (NLP),
% most existing attempts for vision generalists take inspiration from sequence-to-sequence models in the NLP field and model a sequence of discrete tokens through next token prediction.
Pix2Seq \cite{chen2021pix2seq, chen2022pix2} leverages a plain autoregressive transformer and tackles many vision tasks with next-token prediction. For example, bounding boxes in object detection are cast as sequences of discrete tokens, and masks in semantic segmentation are encoded with coordinates of object polygons \cite{castrejon2017annotating}.
The idea was further developed in Unified-IO \cite{lu2022unified}, where dense prediction such as segmentation, depth map, and image restoration are also unified as tokens by using the corresponding image features from a vector quantization variational auto-encoder (VQ-VAE) \cite{van2017vae}.
On the output side, the predicted image tokens are then decoded into masks and depth maps as the final prediction.
% We achieve this unification by homogenizing every supported input and output into a sequence of discrete vocabulary tokens. This common representation across all tasks allows us to train a single transformer-based architecture, jointly on over 90 diverse datasets in the vision and language fields. Unified-IO is the first model capable of performing all 7 tasks on the GRIT benchmark and produces strong results across 16 diverse benchmarks
% unifies object detection, instance segmentation, keypoint detection, and image captioning by quantizing the continuous image coordinates for the first three tasks.
Similarly, OFA \cite{wang2022ofa} unified a diverse set of cross-modal and unimodal tasks in a simple sequence-to-sequence learning framework and achieved competitive performance pretrained with only 20M publicly available image-text pairs.
% OFA \cite{wang2022ofa} provides a direction for unifying different vision tasks using discrete spaces by homogenizing the diverse inputs and outputs to a sequence of discrete tokens.
% Unified-IO \cite{lu2022unified} further unifies dense structure outputs such as images, segmentation masks, and depth maps using a vector quantization variational auto-encoder (VQ-VAE).
Painter \cite{wang2023painter} and SegGPT \cite{wang2023seggpt}, on the other hand, reformulate different vision tasks as an image inpainting problem, and perform in-context learning following \cite{bar2022visual}.
Unlike the previous work, our method unifies different vision tasks under a conditional image generation framework and introduces a diffusion-based vision generalist for it. 

\myparagraph{Unified framework \& Task-specific model:}
\label{sec:related_work}
Besides the aforementioned literature, there is another line of related works that pursue unified architecture but task-specific models. 
UViM \cite{kolesnikov2022uvim} addressed the high-dimensionality output space of vision tasks via learned guiding code, where a short sequence modeled by an additional language model to encode task-specific information guides the prediction of the base model. Separate models are trained for different tasks as the guiding code is task-specific.
XDecoder \cite{zou2023xdec} unified pixel-level image segmentation, image-level retrieval, and vision-language tasks with a generic decoding procedure, which predicts pixel-level masks and token-level semantics, and different combinations of the two outputs are used for different tasks.
Despite their good performance, the task/modality-specific customization poses difficulty for knowledge sharing among different tasks and is also not friendly for supporting unseen tasks.

% unified pixel-labeling tasks with the same modeling approach but trained separate models for different tasks, such as panoptic segmentation, depth estimation and colorization.

% In this work, we explore a diffusion-based vision generalist, where different dense prediction tasks are unified as conditional image generation and we re-purpose pre-trained diffusion models for it.
% Furthermore, we analyze different design choices of diffusion-based generalists and provide a recipe for training such a model.
% In the experiments, we demonstrate the model’s versatility across six different dense prediction tasks and achieve competitive performance to the current state-of-the-art.
% This work, however, is also subject to limitations.
% For example, full fine-tuning of the pre-trained diffusion model at a larger target image resolution can be memory intensive.
% Therefore, the model presented in the paper is trained at a sub-optimal resolution, which limits the final performance across tasks.
% Exploring parameter efficient tuning for such a model would be an interesting future direction.

\section{Toward a Diffusion-Based Generalist} \label{sec:method}
% In this section, we describe our framework.

% \subsection{Preliminaries}
% Diffusion models \cite{???} are a class of generative models that have recently garnered significant attention due to their ability to produce high-quality samples across various domains
% Having been trained on large-scale datasets, they have demonstrated remarkable capabilities of transforming diverse text prompts into realistic images, even for novel concepts.
% In our work, we re-purpose the pretrained text-to-image diffusion models for various dense prediction tasks.

% Diffusion models learn a sequence of state transitions that gradually convert noise, drawn from a predefined noise distribution, into a sample resembling the data distribution p(x).

\begin{figure*}[ht]
\centering
% \vspace{-10pt}
\includegraphics[width=0.9\textwidth]{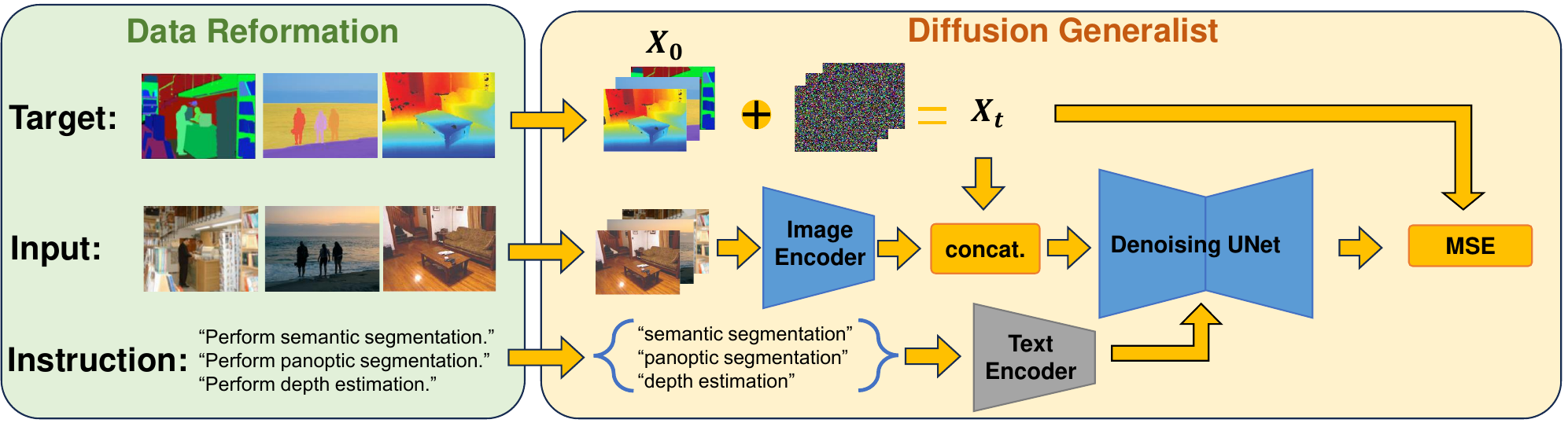}
% \vspace{-5pt}
\caption{
The training pipeline of the diffusion-based vision generalist consists of two parts: \textbf{Left:} Redefining the output space of different vision tasks as RGB images so that they can be unified under a conditional image generation framework. \textbf{Right:} We finetune a pre-trained diffusion model on the reformatted data from the first step. Diffusion is performed in the pixel space to mitigate the quantization error of the latent diffusion (see Table \ref{tab:latent_vs_pixel}). The image and text conditionings are fed into the model via the corresponding encoders, where only the image encoder is tuned during the training.
}
\vspace{-8pt}
\label{fig:model}
\end{figure*}

\subsection{Unification with Conditional Image Generation}
As the output of most vision tasks can be always visualized as images, we redefine the output space of different vision tasks as RGB images and unify them as conditional image generation to tackle the inherent difference of output formats of different vision tasks.
% And then diffusion models can be adapted for it in a multi-task framework.
Given a input image $x$ and the corresponding ground-truth $y$, we first transform $y$ into RGB images and then pair it with a task descriptor in text.
By doing so, training sets of different tasks are combined into a holistic training set. And training the model jointly on it enables the knowledge transfer between tasks.
At test time, given a new image, the model can perform different tasks following the text instructions (examples in Fig. \ref{fig:teaser}).

% consisting of different tasks can be formulated as image generation conditioned on the input image x and the task instruction t.
% Given any conditional image generation model, the model can be trained to solve the tasks contained in the training set.

In this paper, we consider four types of dense prediction tasks: depth estimation, semantic segmentation, panoptic segmentation, and image restoration.

% \myparagraph{Depth estimation}: The original label of the task is the depth value of each pixel ranging from min to max.
% Thus, we map it into 0-255 and then discretize it into integer values, and force the same value across three channels.
\myparagraph{Depth estimation} outputs real number depth value for each pixel on $x$.
Given the minimum and the maximum values, we map them into $[0, 255]$ linearly and discretize them into integers, which is then repeated and stacked along the channel to form the ground-truth RGB label.

\myparagraph{Semantic segmentation} predicts a class label for each pixel. We use a pre-defined injective class-to-color mapping to transform the segmentation mask into RGB images.
Given a task with $C$ categories, we define $C$ colors which are evenly distributed in the 3-dimensional RGB space.
Specifically, following \cite{wang2023painter}, the class index is represented by a 3-digit number with b-base system, where $b=\lceil C^{\frac{1}{3}} \rceil$. 
Thus, the margin between two colors is defined as $\text{int}(\frac{256}{b})$.
The color for the $i$-th class is then [$\text{int}(\frac{i}{b^2}) \times m$, $\text{int}(\frac{i}{b}) \% b \times m$, $l \% b \times m$].
At test time, we find the nearest neighbor of the predicted color in the predefined class-to-color mapping and predict the corresponding category.

\myparagraph{Panoptic segmentation} is solved as a combination of semantic and instance segmentation.
Semantic segmentation labels are constructed as stated above.
For instance segmentation, we set $N$ as the maximum number of instances a single training image can contain.
Then, we define $N$ colors which are evenly distributed in the 3-dimensional RGB space as in semantic segmentation.
% Then, we define instance-to-color mapping
Finally, we assign colors to objects based on their spatial location to form the RGB ground-truth label.
For example, the instance whose center is at the upper leftmost corner obtains the first color and  the lower rightmost gets the last color.
% colors that are evenly distributed in the RGB space and then  
At test time, the model makes predictions twice with different text instructions and merge the results for panoptic segmentation.

\myparagraph{Image restoration} aims to predict the clean image from corrupted images. Thus, the output space is inherently RGB image and does not need further transformation to fit in the framework.

\subsection{A Diffusion Multi-Task Generalist Framework}
By reformating the output space of different vision tasks into images, it is natural to solve them together under a conditional image generation framework.
To this end, we leverage the powerful diffusion models pre-trained for image generation and re-purpose them in our use case.

% As shown in Fig. \ref{fig:model}, the denoising UNet takes the noised target image $x_0$ as the input and generates the output image and the MSE loss is used to train the model.

Fig. \ref{fig:model} shows the overall pipeline of the method, which is a conditional image generation framework with pixel-space diffusion.
Given $M$ tasks with datasets $\{\textbf{I}^i, \textbf{Y}^i\}_{i=1}^M$, where $\textbf{I}^i$ are the input images of task i and $\textbf{Y}^i$ are the corresponding ground-truth labels.
We first transform the output into RGB image format $\textbf{X}^i$ and augment each task with a text instruction $T^i$.
At each training step, we randomly sample a subset of tasks and then sample data from each task.
For each input data $\{I^i, X^i, T^i\}$, we first compute the multi-scale feature map of the original image $I^i$ from the image encoder.
Then, it is concatenated with the noised target image $X_t^i$ before being fed into the UNet for the reconstruction loss.
Note that the image feature can have a different spatial resolution than the target image $X_t^i$, in which case the concatenation will be performed on the interpolated image feature.
In experiments, we find both the image feature resolution and the target resolution are important for the final performance but target resolution matters more.
The text conditioning $T^i$ is fed into the UNet via cross-attention \cite{rombach2022sd}. 
% The input is the noised target image $X_0$ together with the image conditioning and text conditioning.
The whole pipeline is trained in an end-to-end manner except for the text encoder, which is frozen throughout the training.
Compared to the standard diffusion model for conditional image generation, there are three main differences:

\begin{table*}[ht]
\begin{center}
\vspace{-15pt}
\resizebox{0.95\linewidth}{!}{%
\begin{tabular}{rccccccc}
\toprule
\multicolumn{1}{l|}{} & Target & \textbf{Depth Estimation} & \textbf{Semantic Seg.} & \textbf{Panoptic Seg.} & \textbf{Denoising} & \textbf{Deraining} & \textbf{Light Enhance.} \\
\multicolumn{1}{l|}{} & image & RMSE $\downarrow$ & mIoU $\uparrow$ & PQ $\uparrow$ & SSIM $\uparrow$ & SSIM $\uparrow$  & SSIM $\uparrow$ \\
\multicolumn{1}{l|}{} & resolution & NYUv2 & ADE-20K & COCO & SIDD & 5 datasets & LoL \\ \midrule
% \multicolumn{8}{c}{Task-specific models} \\ \midrule
% \multicolumn{1}{r|}{\textcolor{gray}{DenseDepth~\cite{???}}} & & 0.465 & - & - & - & - & -\\
% \multicolumn{1}{r|}{\textcolor{gray}{BinsFormer~\cite{???}}} & & 0.330 & - & - & - & - & -\\
% \multicolumn{1}{r|}{\textcolor{gray}{UperNet-ViT-L~\cite{???}}} & & - & 49.9\% & - & - & - & - \\
% \multicolumn{1}{r|}{\textcolor{gray}{Mask2Former~\cite{???}}} & & - & 57.7\% & 57.8\% & - & - & -\\
% \multicolumn{1}{r|}{\textcolor{gray}{DETR~\cite{???}}} & & - & - & 45.6\% & - & - & -\\
% \multicolumn{1}{r|}{\textcolor{gray}{Uformer~\cite{???}}} & & - & - & - & 0.960 & - & - \\
% \multicolumn{1}{r|}{\textcolor{gray}{MPRNet~\cite{???}}} & & - & - & - & 0.958 & 0.921 & - \\
% \multicolumn{1}{r|}{\textcolor{gray}{MIRNet-v2~\cite{???}}} & & - & - & - & 0.959 & - & 0.851 \\
% \midrule
\multicolumn{8}{c}{Generalist framework, task-specific models} \\ \midrule
\multicolumn{1}{r|}{UViM~\cite{kolesnikov2022uvim}} & $512\times512$ & 0.467 & - & 45.8\% & - & - & - \\ \midrule
\multicolumn{8}{c}{Generalist models} \\ \midrule
\multicolumn{1}{r|}{Unified-IO~\cite{lu2022unified}} & $256\times256$ & 0.385 & 25.7\% & - & - & - & - \\
\multicolumn{1}{r|}{InstructCV~\cite{gan2023instructcv}} & $256\times256$ & \underline{0.297} & 47.2\% & - & - & - & - \\
\multicolumn{1}{r|}{Painter~\cite{wang2023painter}} & $448\times448$ & \textbf{0.288} & \textbf{49.9\%} & \textbf{43.4\%} & \textbf{0.954} & \textbf{0.868} & \textbf{0.872} \\
\multicolumn{1}{r|}{Painter~\cite{wang2023painter}} & $128\times128$ & 0.435\textdagger & 28.4\%\textdagger & 22.6\%\textdagger & 0.922\textdagger & 0.626\textdagger & 0.773\textdagger \\
\multicolumn{1}{r|}{Ours} & $128\times128$ & 0.448 & \underline{48.7\%} & \underline{40.3\%} & \textbf{0.954} & \underline{0.815} & \underline{0.758} \\
\bottomrule
\end{tabular}
}%
\vspace{-5pt}
\caption{
Our method achieves competitive performance in most of the tasks while trained at a much smaller target resolution of $128\times128$. 
When compared at the same resolution, our method shows superior performance over the previous best method (Painter \cite{wang2023painter}), especially on semantic segmentation and panoptic segmentation.
The best number is in bold and the second best number is underscored. \textdagger indicates numbers from our reproduction.
}
\vspace{-20pt}
\label{tab:main_results}
\end{center}
\end{table*}

\begin{table}[ht]
\begin{minipage}{.5\textwidth}
\centering
\includegraphics[width=0.9\textwidth]{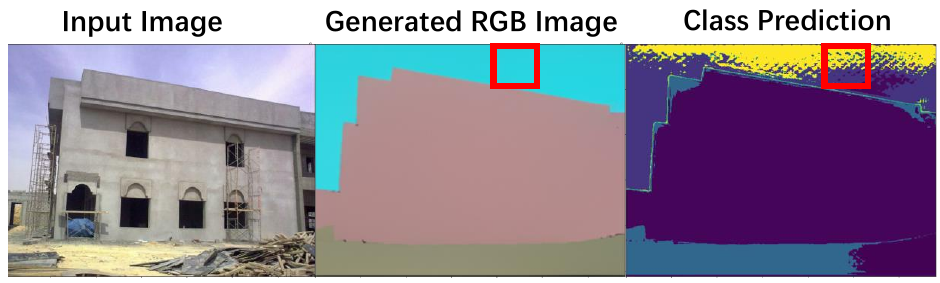}
\end{minipage}

\begin{minipage}{.5\textwidth}
\centering
\resizebox{0.7\linewidth}{!}{%
\begin{tabular}{l|cc}
\toprule
 & \textbf{Semantic Seg.} & \textbf{Panoptic Seg.} \\
 & mIoU $\uparrow$ & PQ $\uparrow$ \\
 & ADE-20K & COCO \\ \midrule
\multicolumn{1}{r|}{Latent Diffusion} & 17.1\% & 11.7\% \\ 
% \multicolumn{1}{r|}{\begin{tabular}[c]{@{}r@{}}Latent Diffusion\\ + VAE Fintuning\end{tabular}} & \multicolumn{1}{l}{} & \multicolumn{1}{l}{} & \multicolumn{1}{l}{} \\ \midrule
\multicolumn{1}{r|}{Pixel Diffusion} & 48.0\% & 35.5\% \\ \bottomrule
\end{tabular}
}%
\end{minipage}%
\vspace{-5pt}
\caption{
\textbf{Upper:}
Semantic segmentation output of the latent diffusion model.
The perceptually same colored regions have different pixel values and, therefore, are mapped to different class labels, leading to bad final performance.
While the red box contains only one ground-truth class sky in generated RGB image, the final class prediction has four classes after the quantization.
\textbf{Lower:}
Latent diffusion suffers from the quantization issue while pixel diffusion achieves good performance.
}
\vspace{-10pt}
\label{tab:latent_vs_pixel}
\end{table}

\begin{itemize}
    \item We propose to directly perform diffusion in the pixel space. 
    As shown in Table \ref{tab:latent_vs_pixel}, when mapping from the latent space to the pixel space, visually uniform regions actually have pixels of many different RGB values.
    This variance can lead to inaccurate class mappings, and consequently, suboptimal performance for semantic and panoptic segmentation.
    % This is because applying the image generation pipeline of latent diffusion models leads to quantization errors (see Table \ref{tab:latent_vs_pixel}), where similar colored regions are mapped to different classes or instances for semantic and panoptic segmentation.

    \item The image conditioning is provided via a feature extractor (we use ConvNeXt \cite{liu2022convnext}) and is concatenated to the target image $X_0$. 
    Compared to the widely adopted method of directly concatenating the raw image as the condition, this brings significant performance improvement, especially for semantic and panoptic segmentation (see Table \ref{tab:diffusion_vs_nondiffusion} for ablation).

    \item We remove the self-attention layers in the outermost layers of UNet. This is because the pixel space diffusion at large target image resolutions induces considerable memory costs. Removing them alleviates the issue without compromising the performance.
\end{itemize}

\section{Experimental Results} \label{sec:experiment}
% In this section, we first highlight important factors in training a diffusion-based multi-task generalist in Section \ref{sec:recipe}. Then, we compare our method with previous approaches in Section \ref{sec:main_results} before the ablation studies in Section \ref{sec:ablation}.
Here, we first explain experimental settings in Section \ref{sec:detail}. Then, we highlight important design choices in diffusion-based multi-task generalists in Section \ref{sec:recipe} before comparing our method with previous approaches in Section \ref{sec:main_results}.

\subsection{Datasets and Implementation Details} \label{sec:detail}

\myparagraph{Datasets:}
We evaluate our method on six different dense prediction tasks with various output formats.
% For depth estimation, we use NYUv2 \cite{???}, which consists of 24K training images and 654 validation images.
% We report the Root Mean Square Error (RMSE).
% For semantic segmentation, we evaluate on ADE20K \cite{???}, which covers a broad range of 150 semantic categories.
% It has 25K images in total, with 20K for training and 2K for validation.
% We adopt the widely used metric of mean IoU (mIoU) for evaluation.
% For panoptic segmentation, we use MS-COCO \cite{???}, which contains approximately 118K training images and 5K validation images with 80 “things” and 53 “stuff” categories. 
% Panoptic segmentation task is evaluated on the union of “things” and “stuff” categories. 
% During inference, the model is forwarded twice for each validation image with different instructions to obtain the results of semantic and instance segmentation respectively.
% The outputs are then merged together to get the results of panoptic segmentation. 
% We report panoptic quality as the measure.
% Image restoration tasks are evaluated on several popular benchmarks, including SIDD \cite{???} for image denoising, LoL \cite{???} for low-light image enhancement, and the merged
% deraining dataset \cite{???} for draining.
For depth estimation, we use NYUv2 \cite{silberman2012nyu} and report the Root Mean Square Error (RMSE).
For semantic segmentation, we evaluate on ADE20K \cite{zhou2017ade} and adopt the widely used metric of mean IoU (mIoU).
For panoptic segmentation, we use MS-COCO \cite{lin2014coco} and report panoptic quality as the measure.
During inference, the model is forwarded twice for each validation image with different instructions to obtain the results of semantic and instance segmentation respectively.
The outputs are then merged together into the panoptic segmentation. 
Image restoration tasks are evaluated on several popular benchmarks, including SIDD \cite{abdelhamed2018sidd} for image denoising, LoL \cite{wei2018lightenhance} for low-light image enhancement, and 5 merged datasets \cite{zamir2022derain} for deraining.

% \myparagraph{Implementation details:}
% As mentioned above, we take the Stable Diffusion v1.4 \cite{???} checkpoint and finetune it jointly on six tasks.
% The image feature extractor is an ImageNet-21K \cite{???} pre-trained ConvNeXt-Large \cite{???}.
% The text encoder is Open-CLIP \cite{???}, which is used in Stable Diffusion \cite{???}.
% We adopt uniform sampling for each tasks except panoptic segmentation, whose weight is twice as much as the other tasks (as it is a combination of semantic and instance segmentation).
% We use AdamW optimizer \cite{???} with constant learning rate of 0.0001, linearly warmed up in the first 20,000 iterations. 
% The target image resolution is 128x128 while the conditioning image resolution is 512x512.
% We train our model for 180,000 steps in total with a batch size of 1024.
\myparagraph{Implementation details.}
As mentioned above, we take the Stable Diffusion v1.4 \cite{rombach2022sd} checkpoint and finetune it jointly on six tasks.
The image feature extractor is an ImageNet-21K \cite{russakovsky2015imagenet} pre-trained ConvNeXt-Large \cite{liu2022convnext}.
The text encoder is Open-CLIP \cite{radford2021clip}, which is used in Stable Diffusion \cite{rombach2022sd}.
We adopt uniform sampling for each tasks except panoptic segmentation, whose weight is twice as much as the other tasks (as it is a combination of semantic and instance segmentation).
Following \cite{chen2023bits}, we also adjust the input scaling factor by a constant factor $b$ in the forward noising processing of diffusion.
We use AdamW optimizer \cite{kingma2014adam} with constant learning rate of 0.0001, linearly warmed up in the first 20,000 iterations. 
The target image resolution is $128\times128$ while the conditioning image resolution is $512\times512$.
We train our model for 180,000 steps in total with a batch size of 1024.

\begin{table*}[ht]
\begin{center}
% \vspace{-15pt}
\resizebox{0.8\linewidth}{!}{%
\begin{tabular}{rcccccc}
\toprule
\multicolumn{1}{l|}{} & \textbf{Depth Estimation} & \textbf{Semantic Seg.} & \textbf{Panoptic Seg.} & \textbf{Denoising} & \textbf{Deraining} & \textbf{Light Enhance.} \\
\multicolumn{1}{l|}{} & RMSE $\downarrow$ & mIoU $\uparrow$ & PQ $\uparrow$ & SSIM $\uparrow$ & SSIM $\uparrow$  & SSIM $\uparrow$\\
\multicolumn{1}{l|}{} & NYUv2 & ADE-20K & COCO & SIDD & 5 datasets & LoL \\ \midrule
\multicolumn{1}{r|}{Ours} & 0.511 & \textbf{48.0\%} & \textbf{35.5\%} & 0.949 & 0.772 & \textbf{0.704} \\ \midrule
\multicolumn{1}{r|}{Non-diffusion} & \textbf{0.443} & 42.4\% & 19.8\% & \textbf{0.951} & \textbf{0.773} & 0.703 \\
\multicolumn{1}{r|}{Train from scratch} & 0.528 & 46.6\% & 33.6\% & 0.948 & 0.764 & \textbf{0.704 }\\ 
\multicolumn{1}{r|}{Direct concat.} & 0.476 & 37.6\% & 27.1\% & 0.941 & 0.772 & 0.687 \\
\bottomrule
\end{tabular}
}%
% \vspace{-5pt}
\caption{
We analyze the important design choices of our method and aim to provide a recipe for training diffusion-based generalists: 1. diffusion models greatly outperform non-diffusion models on panoptic segmentation; 2. text-to-image generation pre-training leads to an overall better performance; 3. conditioning on image features extracted from an encoder gives significant improvement over the raw image.
}
\vspace{-10pt}
\label{tab:diffusion_vs_nondiffusion}
\end{center}
\end{table*}

\begin{figure*}[ht]
    \centering
    \includegraphics[width=\linewidth]{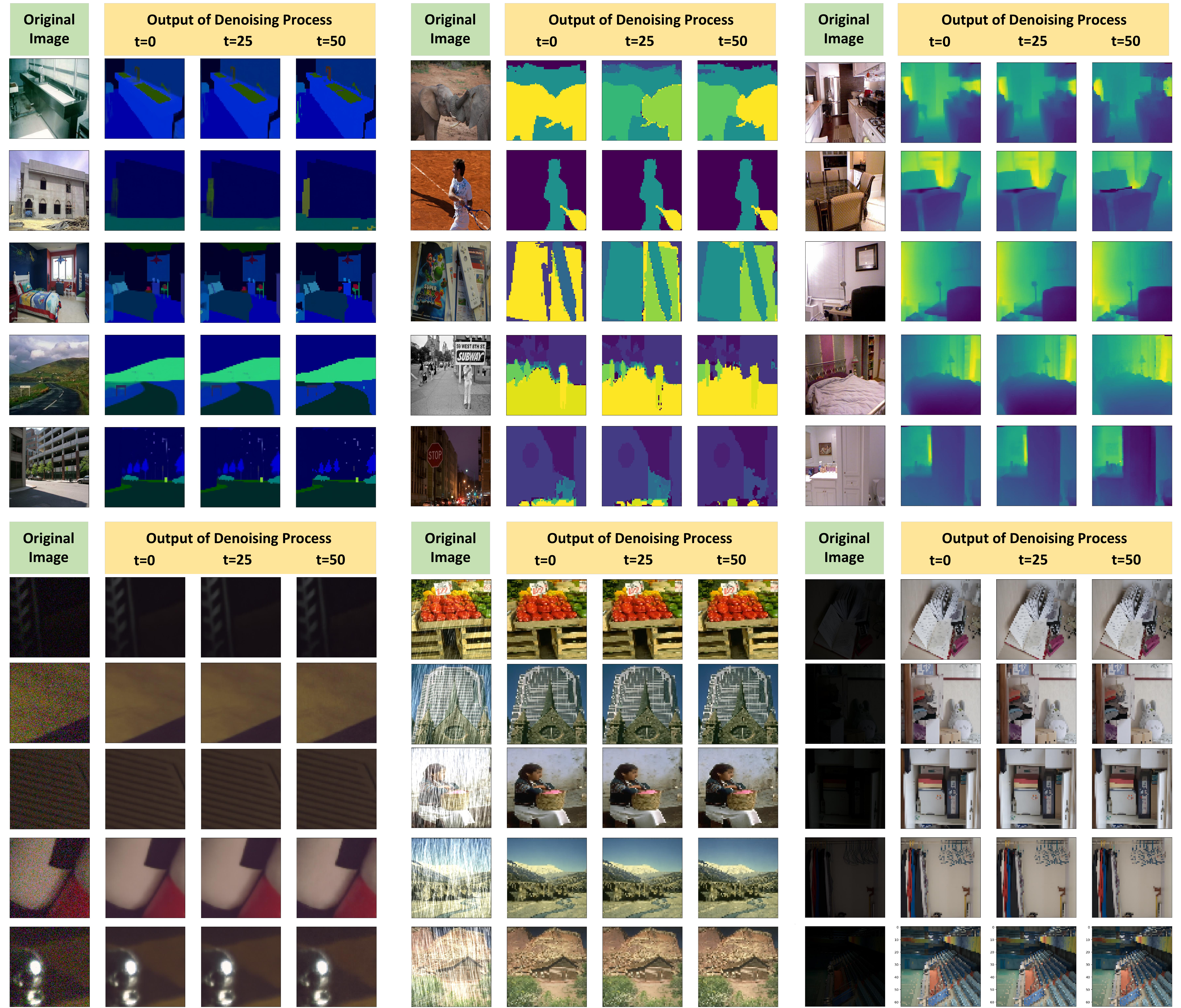}
    \caption{Qualitative results on images from the validation sets of ADE20K, MS-COCO, NYU-V2, SIDD, Deraining, and LOL. Following a raster scan order, the text prompts are "Performance semantic segmentation", "Performance instance segmentation", "Performance depth estimation", "Performance image restoration denoising", "Performance image restoration deraining", and "Performance image restoration light enhancement", respectively. The images are not cherry-picked.}
    \label{fig_visual:1}
\end{figure*}

\subsection{Recipes for Diffusion-Based Generalists} \label{sec:recipe}
In this section, we analyze the design choices of our method and show their importance through ablation experiments. 
Specifically, we show the importance of diffusion by training the same model as in Fig. \ref{fig:model} to directly generate target images without using diffusion (non-diffusion). 
We study the significance of image generation pre-training and image encoder by training models without them (train from scratch and direct concat.).
If not specified, we train all models at a target resolution of $64\times64$ for 50,000 steps. % for fair comparison.

We attribute the success of our method to four aspects. 
(1) While having similar results on image restoration tasks, diffusion-based generalist achieves better performance than non-diffusion models on segmentation tasks which requires a global understanding of the scene and the semantics.
For example, the diffusion model reaches 35.5\% PQ for panoptic segmentation while the non-diffusion model has only 19.8\% (Table \ref{tab:diffusion_vs_nondiffusion} ours v.s. non-diffusion).
(2) Image generation pre-training on large scale dataset transfers useful knowledge to the many downstream tasks.
The model finetuned from Stable Diffusion v1.4 \cite{rombach2022sd} achieves better results than the one trained from scratch across the tasks (Table \ref{tab:diffusion_vs_nondiffusion} ours vs train from scratch).
(3) The image conditioning can take advantage of powerful pre-trained image encoders by conditioning on the image features rather than the raw image, which is in contrast to the standard practice for image generation tasks.
On semantic segmentation and panoptic segmentation, extracting features gives 10.4\% and 8.4\% performance improvement, respectively (Table \ref{tab:diffusion_vs_nondiffusion} ours v.s. direct concat.).
(4) Pixel diffusion is better than latent diffusion as it does not suffer from the quantization issue while upsampling (see Table \ref{tab:latent_vs_pixel} for an example).

% \myparagraph{Image generation pretraining stabilizes the training.}

% \begin{table}[ht]
% \begin{center}
% \resizebox{0.97\linewidth}{!}{%
% \begin{tabular}{l|cccccc}
% \toprule
%  & \textbf{Depth Estimation} & \textbf{Semantic Seg.} & \textbf{Panoptic Seg.} & \textbf{Denoising} & \textbf{Deraining} & \textbf{Light Enhance.} \\
%  & RMSE $\downarrow$ & mIoU $\uparrow$ & PQ $\uparrow$ & SSIM $\uparrow$ & SSIM $\uparrow$  & SSIM $\uparrow$ \\
%  & NYUv2 & ADE-20K & COCO & SIDD & 5 datasets & LoL \\ \midrule
% \multicolumn{1}{r|}{Train from scratch} & 0.595 & 47.2\% & 38.9\% \\
% \multicolumn{1}{r|}{Stable Diffusion ckpt} & 0.419 & 50.4\% & 41.2\% \\ \bottomrule
% \end{tabular}
% }%
% % \vspace{-5pt}
% \caption{
% Image generation pretraining stabilizes the training.
% }
% \label{tab:sd_ckpt}
% \end{center}
% \end{table}

\subsection{Comparisons with Prior Art} \label{sec:main_results}
We compare our model with recent vision generalists in Table \ref{tab:main_results}.
With a much smaller target image resolution at $128\times128$, our method achieves competitive performance across the tasks.
In particular, when compared with the previous best model Painter \cite{wang2023painter} at the same target resolution, our method has a significant margin over them, which highlights the potential of our method at a higher resolution.

\subsection{Qualitative Results}
In this section, we visualize the output of our method on six different tasks in Fig. \ref{fig_visual:1}.
We use DDIM at inference time with 50 steps.
Each figure shows the output of the denoising process at the 0-th, 25-th, and 50-th steps.

\subsection{Ablation Study}  \label{sec:ablation}
In this section, we analyze the effect of other important hyper-parameters of our method, such as batch size, target image resolution, and noise-signal ratio. %, and joint training versus separate training.
Similar to Section \ref{sec:recipe}, we train all models at a target resolution of $64\times64$ for 50,000 steps by default. % for fair comparison.

\myparagraph{Effect of batch size.}
Here, we discuss the effect of different batch sizes for our method. 
As shown in Table \ref{tab:bs}, the performance of most of the tasks improves with the increase of the batch size. In particular, panoptic segmentation greatly benefits from the large batch size.
% \begin{table}[ht]
% \begin{center}
% \resizebox{0.97\linewidth}{!}{%
% \begin{tabular}{r|ccc}
% \toprule
% \multicolumn{1}{c|}{\multirow{3}{*}{Batch size}} & \textbf{Depth Estimation} & \textbf{Semantic Seg.} & \textbf{Panoptic Seg.} \\
% \multicolumn{1}{c|}{} & RMSE $\downarrow$ & mIoU $\uparrow$ & PQ $\uparrow$ \\
% \multicolumn{1}{c|}{} & NYUv2 & ADE-20K & COCO \\ \midrule
% 128 &  &  &  \\
% 256 &  &  &  \\
% 512 &  &  &  \\
% 1024 &  &  &  \\ 
% \bottomrule
% \end{tabular}
% }%
% % \vspace{-5pt}
% \caption{
% Large batch size improves the performance by mitigating underfitting.
% }
% \label{tab:bs}
% \end{center}
% \end{table}

\begin{table}[ht]
\begin{center}
\resizebox{1\linewidth}{!}{%
\setlength\tabcolsep{3pt}
\begin{tabular}{r|cccccc}
\toprule
\multicolumn{1}{l|}{} & \textbf{Depth} & \textbf{Sem. Seg.} & \textbf{Pan. Seg.} & \textbf{Denoise} & \textbf{Detrain} & \textbf{Enhance.} \\
\multicolumn{1}{l|}{} & RMSE $\downarrow$ & mIoU $\uparrow$ & PQ $\uparrow$ & SSIM $\uparrow$ & SSIM $\uparrow$  & SSIM $\uparrow$\\
\multicolumn{1}{l|}{} & NYUv2 & ADE-20K & COCO & SIDD & 5 datasets & LoL \\ \midrule
128 & 0.548 & 35.5\% & 26.2\% & 0.941 & 0.754 & 0.701 \\
256 & 0.495 & 44.3\% & 30.0\% & 0.945 & 0.766 & 0.703 \\
512 & \textbf{0.491} & 47.1\% & 33.5\% & 0.948 & 0.770 & 0.702 \\
1024 & 0.511 & \textbf{48.0\%} & \textbf{35.5\%} & \textbf{0.949} & \textbf{0.772} & \textbf{0.704} \\
\bottomrule
\end{tabular}
}%
\vspace{-5pt}
\caption{
Large batch size improves the performance for all the tasks except depth estimation.
}
\label{tab:bs}
\vspace{-20pt}
\end{center}
\end{table}

% \myparagraph{Effect of different image conditioning.}

% \begin{table}[ht]
% \begin{center}
% \resizebox{0.97\linewidth}{!}{%
% \begin{tabular}{r|ccc}
% \toprule
% \multicolumn{1}{c|}{\multirow{3}{*}{\begin{tabular}[c]{@{}c@{}}Image\\  Conditioning\end{tabular}}} & \textbf{Depth Estimation} & \textbf{Semantic Seg.} & \textbf{Panoptic Seg.} \\
% \multicolumn{1}{c|}{} & RMSE $\downarrow$ & mIoU $\uparrow$ & PQ $\uparrow$ \\
% \multicolumn{1}{c|}{} & NYUv2 & ADE-20K & COCO \\ \midrule
% Cross-attention & 0.491 & 49.8\% & 40.0\% \\
% Concat. & 0.419 & 50.4\% & 41.2\% \\ \bottomrule
% \end{tabular}
% }%
% % \vspace{-5pt}
% \caption{
% Effect of different image conditioning.
% }
% \label{tab:image_cond}
% \end{center}
% \end{table}

\myparagraph{Effect of target resolution.}
Table \ref{tab:mask_resolution} studies the effect of different target image resolutions.
Since our method performs diffusion in the pixel space, increasing the target image resolution is important for good performance.
Despite the increased memory cost, our method achieves its best performance at the resolution of $128\times128$ and can be further improved with even larger target images.
% \begin{table}[ht]
% \begin{center}
% \resizebox{0.97\linewidth}{!}{%
% \begin{tabular}{r|ccc}
% \toprule
% \multicolumn{1}{c|}{\multirow{3}{*}{\begin{tabular}[c]{@{}c@{}}Output\\  Resolution\end{tabular}}} & \textbf{Depth Estimation} & \textbf{Semantic Seg.} & \textbf{Panoptic Seg.} \\
% \multicolumn{1}{c|}{} & RMSE $\downarrow$ & mIoU $\uparrow$ & PQ $\uparrow$ \\
% \multicolumn{1}{c|}{} & NYUv2 & ADE-20K & COCO \\ \midrule
% 32x32 &  &  &  \\
% 64x64 &  &  &  \\
% 128x128 & 0.419 & 50.4\% & 41.2\% \\
% % 256x256 &  &  &  \\ 
% \bottomrule
% \end{tabular}
% }%
% % \vspace{-5pt}
% \caption{
% Effect of output resolution.
% }
% \label{tab:mask_resolution}
% \end{center}
% \end{table}

\begin{table}[ht]
% \vspace{-5pt}
\begin{center}
\resizebox{1\linewidth}{!}{%
\setlength\tabcolsep{3pt}
\begin{tabular}{r|cccccc}
\toprule
\multicolumn{1}{l|}{} & \textbf{Depth} & \textbf{Sem. Seg.} & \textbf{Pan. Seg.} & \textbf{Denoise} & \textbf{Detrain} & \textbf{Enhance.} \\
\multicolumn{1}{l|}{} & RMSE $\downarrow$ & mIoU $\uparrow$ & PQ $\uparrow$ & SSIM $\uparrow$ & SSIM $\uparrow$  & SSIM $\uparrow$\\
\multicolumn{1}{l|}{} & NYUv2 & ADE-20K & COCO & SIDD & 5 datasets & LoL \\ \midrule
32x32 & 0.514 & 44.4\% & 32.1\% & 0.940 & 0.743 & 0.653 \\
64x64 & 0.511 & 48.0\% & 35.5\% & 0.949 & 0.772 & 0.704 \\
128x128 & \textbf{0.467} & \textbf{49.2\%} & \textbf{36.7\%} & \textbf{0.953} & \textbf{0.810} & \textbf{0.762} \\
\bottomrule
\end{tabular}
}%
\vspace{-5pt}
\caption{
Effect of output resolution. Increasing the target image resolution significantly improves the performance across tasks.
}
\vspace{-20pt}
\label{tab:mask_resolution}
\end{center}
\end{table}

% \myparagraph{Additional downsampling improves efficiency.}

% \begin{table}[ht]
% \begin{center}
% \resizebox{0.97\linewidth}{!}{%
% \begin{tabular}{rll}
% \toprule
%  & w/ auxiliary down/up & w/o auxiliary down/up \\ \midrule
% \begin{tabular}[c]{@{}r@{}}Depth Estimation\\ RMSE $\downarrow$ NYUv2\end{tabular} &  &  \\
% \begin{tabular}[c]{@{}r@{}}Semantic Seg.\\ mIoU $\uparrow$ ADE20-K\end{tabular} &  &  \\
% \begin{tabular}[c]{@{}r@{}}Panoptic Seg.\\ PQ $\uparrow$ COCO\end{tabular} &  &  \\ \midrule
% FLOPs &  &  \\
% Memory cost &  &  \\ \bottomrule
% \end{tabular}
% }%
% % \vspace{-5pt}
% \caption{
% Additional downsampling improves efficiency.
% }
% \label{tab:aux_down}
% \end{center}
% \end{table}

\myparagraph{Importance of noise-signal ratio.}
In DDPM \cite{ho2020ddpm}, the forward diffusion process is defined as $x_t = \sqrt{\gamma_t}x_0+\sqrt{1-\gamma_t}\epsilon$, where
$x_0$ is the input image, $\epsilon$ is a Gaussian noise, and $t$ is the number of diffusion step.
As shown in \cite{chen2023bits}, the denoising task at the same noise level (i.e. the same t) becomes simpler with the increase in the image size.
In order to compensate for this, \cite{chen2023bits} proposed to scale the input with a constant $b$ to explicitly control the noise-signal ratio, which results in the forward diffusion process as $x_t = \sqrt{\gamma_t}bx_0+\sqrt{1-\gamma_t}\epsilon$.
As we reduce $b$, it increases the noise levels.
Table \ref{tab:nsr} shows the effect of the noise-signal ratio $b$ where $b=0.5$ gives the best performance.
% \begin{table}[ht]
% \begin{center}
% \resizebox{0.97\linewidth}{!}{%
% \begin{tabular}{r|ccc}
% \toprule
% \multicolumn{1}{c|}{\multirow{3}{*}{\begin{tabular}[c]{@{}c@{}}Noise-signal\\  Ratio\end{tabular}}} & \textbf{Depth Estimation} & \textbf{Semantic Seg.} & \textbf{Panoptic Seg.} \\
% \multicolumn{1}{c|}{} & RMSE $\downarrow$ & mIoU $\uparrow$ & PQ $\uparrow$ \\
% \multicolumn{1}{c|}{} & NYUv2 & ADE-20K & COCO \\ \midrule
% 0.1 &  &  &  \\
% 0.3 & 0.419 & 50.4\% & 41.2\% \\
% 0.5 &  &  &  \\
% 0.7 &  &  &  \\
% 1.0 &  &  &  \\ \bottomrule
% \end{tabular}
% }%
% % \vspace{-5pt}
% \caption{
% Importance of noise-signal ratio.
% }
% \label{tab:bs}
% \end{center}
% \end{table}

\begin{table}[ht]
\begin{center}
\resizebox{1\linewidth}{!}{%
\setlength\tabcolsep{3pt}
\begin{tabular}{r|cccccc}
\toprule
\multicolumn{1}{l|}{} & \textbf{Depth} & \textbf{Sem. Seg.} & \textbf{Pan. Seg.} & \textbf{Denoise} & \textbf{Detrain} & \textbf{Enhance.} \\
\multicolumn{1}{l|}{} & RMSE $\downarrow$ & mIoU $\uparrow$ & PQ $\uparrow$ & SSIM $\uparrow$ & SSIM $\uparrow$  & SSIM $\uparrow$\\
\multicolumn{1}{l|}{} & NYUv2 & ADE-20K & COCO & SIDD & 5 datasets & LoL \\ \midrule
0.1 & \textbf{0.497} & 46.9\% & 33.1\% & 0.948 & 0.770 & 0.702 \\
0.3 & 0.511 & {48.0\%} & {35.5\%} & \textbf{0.949} & {0.772} & {0.704} \\
0.5 & 0.514 & \textbf{49.3\%} & \textbf{35.9\%} & \textbf{0.949} & \textbf{0.774} & \textbf{0.708} \\
0.7 & 0.533 & 48.2\% & 34.4\% & \textbf{0.949} & 0.773 & 0.707 \\
1.0 & 0.572 & 40.3\% & 31.1\% & 0.948 & 0.770 & 0.706 \\
\bottomrule
\end{tabular}
}%
\vspace{-5pt}
\caption{
Importance of noise-signal ratio $b$ in the forward diffusion process $x_t = \sqrt{\gamma_t}bx_0+\sqrt{1-\gamma_t}\epsilon$.
}
\vspace{-20pt}
\label{tab:nsr}
\end{center}
\end{table}

\section{Conclusion and Limitations} \label{sec:conclusion}
In this work, we explore a diffusion-based vision generalist, where different dense prediction tasks are unified as conditional image generation and we re-purpose pre-trained diffusion models for it.
Furthermore, we analyze different design choices of diffusion-based generalists and provide a recipe for training such a model.
In experiments, we demonstrate the model’s versatility across six different dense prediction tasks and achieve competitive performance to the current state-of-the-art.
This work, however, is also subject to limitations.
% For example, full fine-tuning of the pre-trained diffusion model at a larger target image resolution can be memory intensive.
For example, full fine-tuning of the pre-trained diffusion model at a larger target image resolution is memory intensive due to the pixel space diffusion.
% Thus, the model presented in the paper is trained at a sub-optimal resolution, which limits the final performance across tasks.
Thus, exploring parameter-efficient tuning for such a model would be an interesting future direction.

% Secondly, the current method is not applicable in the open-vocabulary setting, which can be an interesting future direction as well.

{
    \small
    \bibliographystyle{ieeenat_fullname}
    \bibliography{main}
}

\end{document}